\documentclass[letterpaper, 10 pt, conference]{ieeeconf}  

\IEEEoverridecommandlockouts                              

\usepackage{color}
\usepackage{graphicx}
\usepackage{float}
\usepackage{amsmath}
\usepackage{adjustbox}

\usepackage{soul}
\usepackage{verbatim}

\title{\LARGE \bf
A User Study Method on Healthy Participants for Assessing an Assistive Wearable Robot Utilising EMG Sensing
}

\author{Cem Suulker, Alexander Greenway, Sophie Skach, Ildar Farkhatdinov, Stuart Charles Miller, and Kaspar Althoefer
\thanks{Full version of this extended abstract is under review for IEEE RA-L journal.}
\thanks{All authors are with the Centre for Advanced Robotics @ Queen Mary, School of Engineering and Materials Science, Queen Mary University of London, United Kingdom.
        {\tt\footnotesize c.suulker@qmul.ac.uk}}%
}

\begin{document}

\maketitle
\thispagestyle{empty}
\pagestyle{empty}

\begin{abstract}

Hand-wearable robots, specifically exoskeletons, are designed to aid hands in daily activities, playing a crucial role in post-stroke rehabilitation and assisting the elderly. Our contribution to this field is a textile robotic glove with integrated actuators. These actuators, powered by pneumatic pressure, guide the user's hand to a desired position. Crafted from textile materials, our soft robotic glove prioritizes safety, lightweight construction, and user comfort. Utilizing the ruffles technique, integrated actuators guarantee high performance in blocking force and bending effectiveness. Here, we present a participant study confirming the effectiveness of our robotic device on a healthy participant group, exploiting EMG sensing.

\end{abstract}

\section{Introduction}

Human reliance on finger dexterity is paramount for accomplishing everyday tasks involving pinching or grasping various objects. However, individuals affected by neurogenerative disorders, such as strokes or Parkinson's disease, may experience limitations in dexterity levels. In response to this challenge, wearable robotic devices have emerged to assist individuals with such tasks, showcasing increasing success \cite{jiryaei2024usefulness}. Beyond addressing specific grasping objectives, hand-wearable robots have the potential to offer broader benefits, thereby augmenting human capabilities. These advantages extend to activities such as lifting heavy objects (e.g., moving furniture) and are particularly applicable in industrial settings where human mobility is crucial.

By employing actuators, one can direct fingers into a pinch grasp position and subsequently apply forces to the target object. Extensive research has focused on such actuators, with recent emphasis shifting towards soft robotic variants compared to their rigid counterparts \cite{zhu2022soft}. Enhanced safety and comfort for users have been achieved through the increased utilization of soft materials. Notably, pneumatically driven elastomers and textile structures have demonstrated reliability in achieving necessary blocking force and finger bending for routine daily tasks \cite{suulker2022comparison}. This shift is promising, as wearable soft robotic structures offer superior convenience compared to rigid links and rotary motors due to their lighter weight and safer nature.

\begin{figure}[t]
  \centering
  \includegraphics[width=1\linewidth]{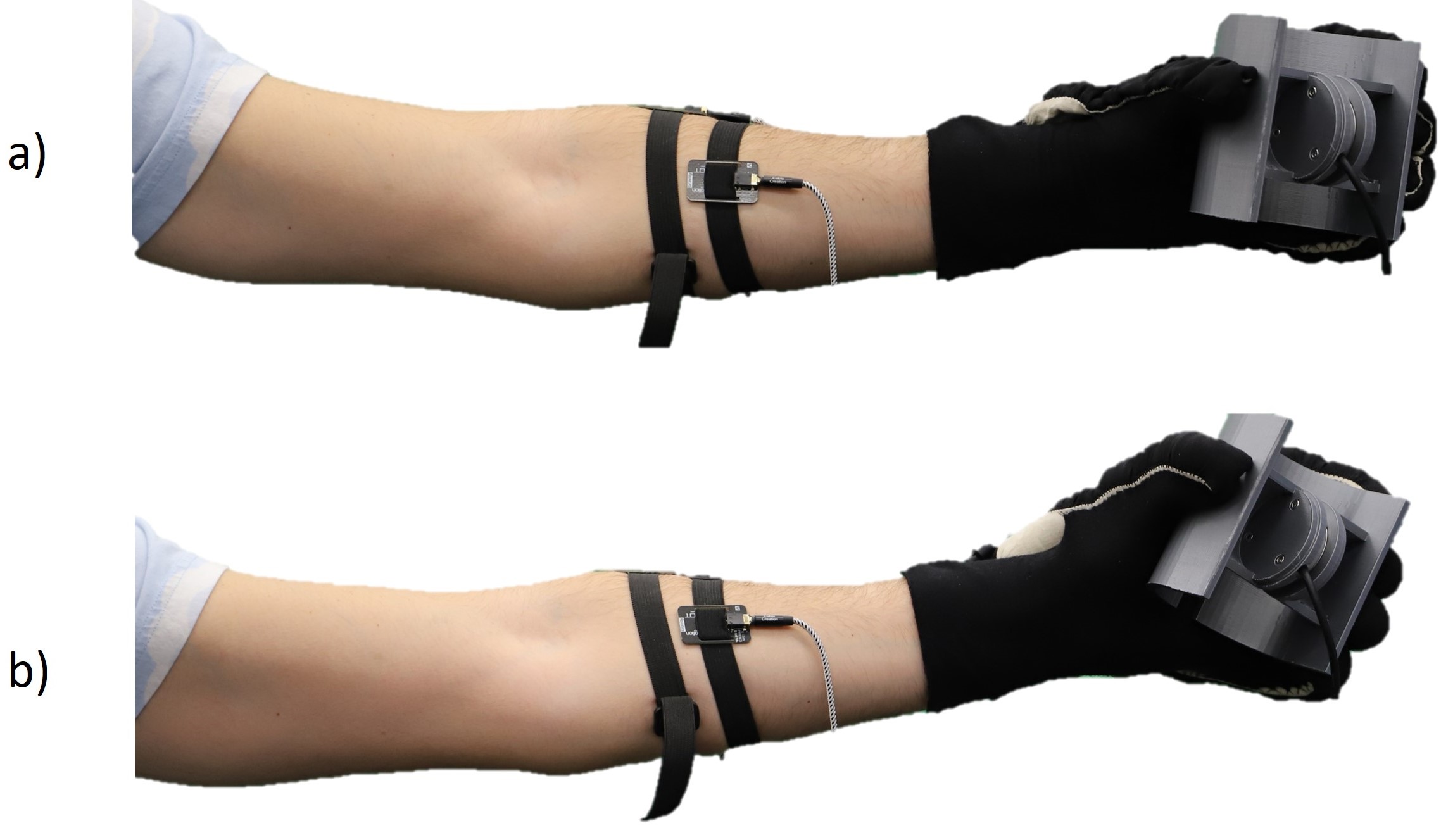}
  \caption{The arm of a participant when grasping a force sensor integrated object. Using two sEMG sensors, we monitor how much muscle activity is involved a) when the assitive device is off, b) when the assistive device is on.}
  \label{fig1}
\end{figure}

Established methodologies exist in the literature for evaluating the effectiveness of hand-wearable assistive devices based on data from user studies. For instance, the Jebsen Taylor Hand Function Test \cite{jebsen} and the Box and Blocks Test \cite{mathiowetz1985adult} are commonly employed to assess the benefits of assistive devices \cite{correia2020improving}. While these tests prove effective with individuals exhibiting compromised manual dexterity, significant challenges arise when working with such participant groups. Apart from ensuring the safety of robotic prototypes for fragile or vulnerable individuals, complex ethical issues must be addressed. Many researchers, therefore, prefer working with healthy sample groups. However, this introduces complexities, particularly regarding trust, as the natural dexterity of healthy users can influence test results by either aiding or interfering with robotic device actuation. Given the high flexibility and relatively low actuation force of soft robotic devices, utilizing a healthy sample for testing is not advisable. Instead, Electromyography (EMG) sensing should be considered to quantify the assistance provided by the robotic device.

Surface EMG (sEMG) sensors serve as valuable tools for measuring electrical signals generated by muscles during contraction, facilitating insights into muscle activity for various applications in biomechanical research \cite{taborri2020sport}, rehabilitation \cite{delph2013soft,guo2018soft}, and the development of prosthetics \cite{khushaba2012toward} or assistive devices \cite{zhou2019soft}. In the context of this paper, we focus on using EMG signals to evaluate the effectiveness of an assistive device \cite{young2023air}. By measuring the magnitude of the differential between signals at rest and under maximum voluntary contraction (MVC), EMG signals can quantify muscle activity with and without the assistive device, thus assessing its effectiveness.

This paper introduces a novel method validated by sEMG sensors to evaluate the effectiveness of a wearable assistive hand device (Fig. \ref{fig1}). We then assess our soft robotic assistive prototype glove using this method and demonstrate its efficacy in assisting individuals in pinch grasping objects. EMG signals are recorded as participants perform a task involving varying levels of pinch grasp force on an object equipped with a force sensor. Our findings shows significant results, validated using linear mixed effect models.

\section{The Assistive Glove} 

To create bending of an inflatable textile actuator, an imbalance between two layers of fabric must be achieved. This imbalance is created by using pleating techniques \cite{cappello2018assisting}, excess material of one layer is folded and stitched onto the other, and unfolds when inflated. This approach, however, has potential drawbacks, e.g. when the space between the layers is too small for the pleats to unfold (which, for finger sized designs, can become imminent). Another technique to create such imbalance is the use of different types of fabric with varying elasticity. This way, one layer stretches more when actuated and the structure will bend towards the less elastic one. Integrating braided elastic band to the stretch fabric option with the ruffles technique is proven more efficient in terms of blocking force and bending angle capabilities \cite{suulker2022soft}.

\begin{figure}[h]
  \centering
  \includegraphics[width=1\linewidth]{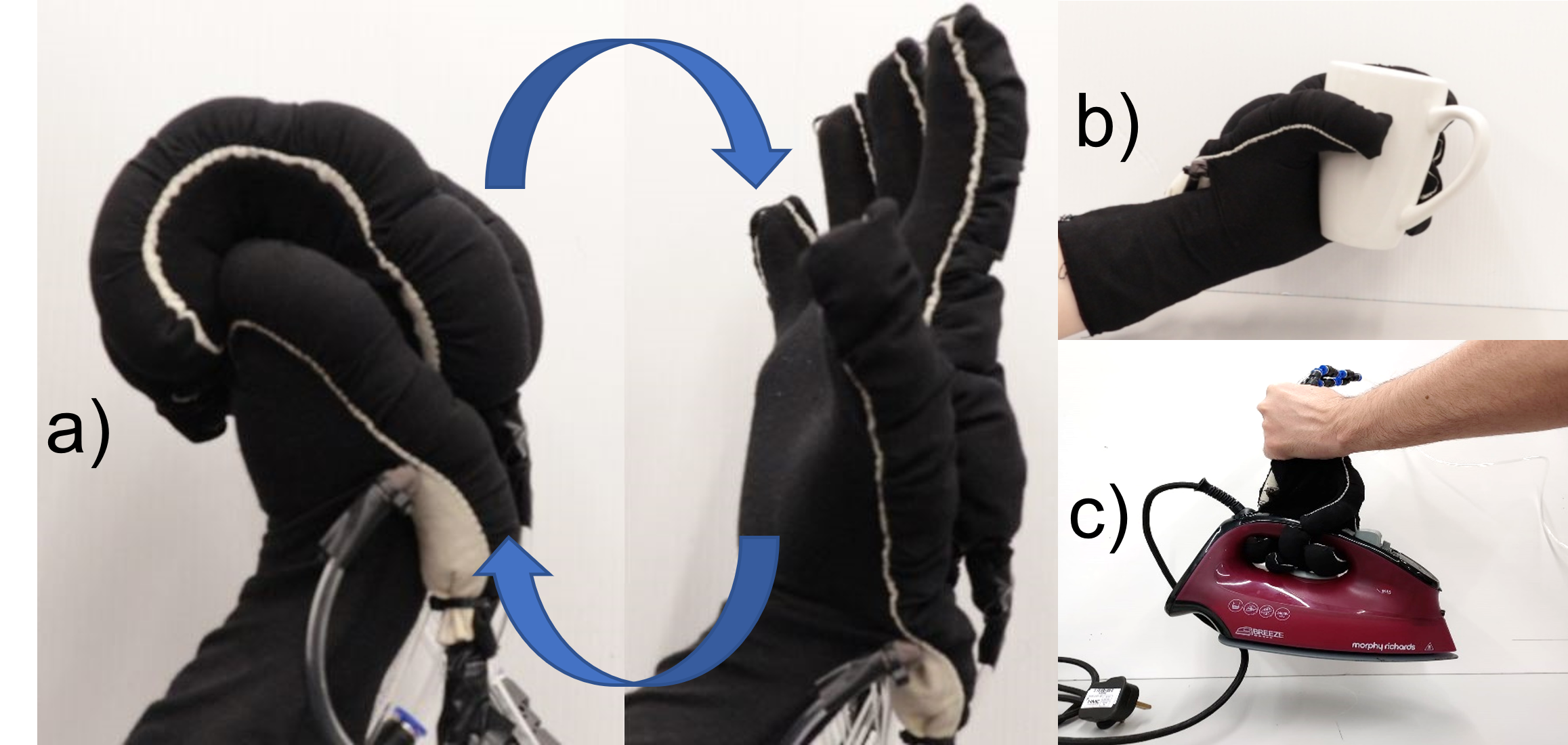}
  \caption{The soft robotic glove prototype in the action of a) closing and opening of the hand, b) grasp assistance, c) lifting an iron (1.3 kg) without a hand in it.}
  \label{fig2}
\end{figure}

\subsubsection{Materials} 

To create this textile actuator the two different textile properties layers are selected. Bottom: a plain cotton weave for the (light fabric in Fig. \ref{fig2}). Top: a cotton mix with elastane yarn integrated to the weft of the fabric (dark fabric in Fig. \ref{fig2}).

To enhance a fabric's stretch behavior and create the desired material imbalance between layers, an additional support material is used that is integrated when assembling the actuators: a braided elastic band, also called elastics, commonly used in clothing to create ruffles or elastic waistbands. It consists of braided polyester and a small part of thin rubber, making it durable and extremely stretchy.

\subsubsection{Fabrication}

First, the mono-directional stretch black fabric was cut 80\% longer than the cotton bottom layer fabric. The elastic band was integrated on the side seam between the top and bottom layer, first stitched onto the top layer while being stretched, and then sewn onto the bottom layer in a relaxed state. This enabled the top layer to reach an excess length of 180\% of the bottom layer. The actuator is equipped with a latex bladder to ensure air tightness.

\begin{figure}[h]
\centering
  \includegraphics[width=1\linewidth]{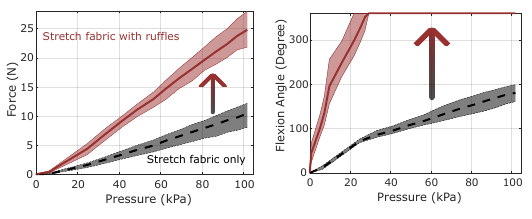}
  \caption{Blocking force output and flexion angle versus pressure graphs for stretch fabric actuator, and elastic band integrated actuator. Integration of the elastic band significantly boosts the performance.}
  \label{fig3}
\end{figure}

\subsubsection{Results}
Two important parameters for soft bending actuators for wearables are flexion angle and blocking force \cite{suulker2022fabric}. In example, for rehabilitation or assistive hand exoskeletons it is imperative that each finger is unrestricted in relation to its maximum angle, and actuators should apply 10-15 N blocking force (Fig. \ref{fig3}) to the fingers\cite{takahashi2008robot}. 

In Fig. 2 the elastic band integration boosts both of these critical parameters for the actuator. The force capability increases approximately from 10 N to 25 N, and the maximum bending angle increases from approximately 180 degrees to 360 degrees (Fig. \ref{fig3}).

\section{Assessment of the Assistive Glove}


The existing literature offers established methodologies for evaluating the effectiveness of hand-wearable assistive devices through user studies, utilizing tests such as the Jebsen Taylor Hand Function Test \cite{jebsen} and the Box and Blocks Test \cite{mathiowetz1985adult}. These assessments are particularly valuable when conducted with individuals whose manual dexterity is compromised in some capacity. However, there are considerable challenges associated with working with such participant groups. In addition to ensuring the safety of the robotic prototype for use by fragile or vulnerable individuals, complex ethical considerations must be addressed.

Given these challenges, many researchers opt to conduct studies with sample groups comprised of relatively "unproblematic" individuals \cite{maldonado2023fabric}. Nevertheless, this approach introduces its own complexities, particularly concerning trust. Healthy or non-compromised users may inadvertently influence test results due to their natural dexterity. Their ability to grasp and hold objects can either override or assist the actuation of the robotic device. This issue is particularly pertinent with soft robotic devices, which possess high flexibility and relatively low actuation force. Consequently, it is not advisable to work with a sample of healthy or non-compromised individuals.

Instead, to accurately quantify the assistance provided by the robotic device, Electromyography sensing should be considered. EMG sensing offers a more objective measure of the user's interaction with the device, allowing researchers to evaluate the device's effectiveness in providing assistance independently of the user's inherent dexterity.

\begin{figure}[h]
\centering
  \includegraphics[width=1\linewidth]{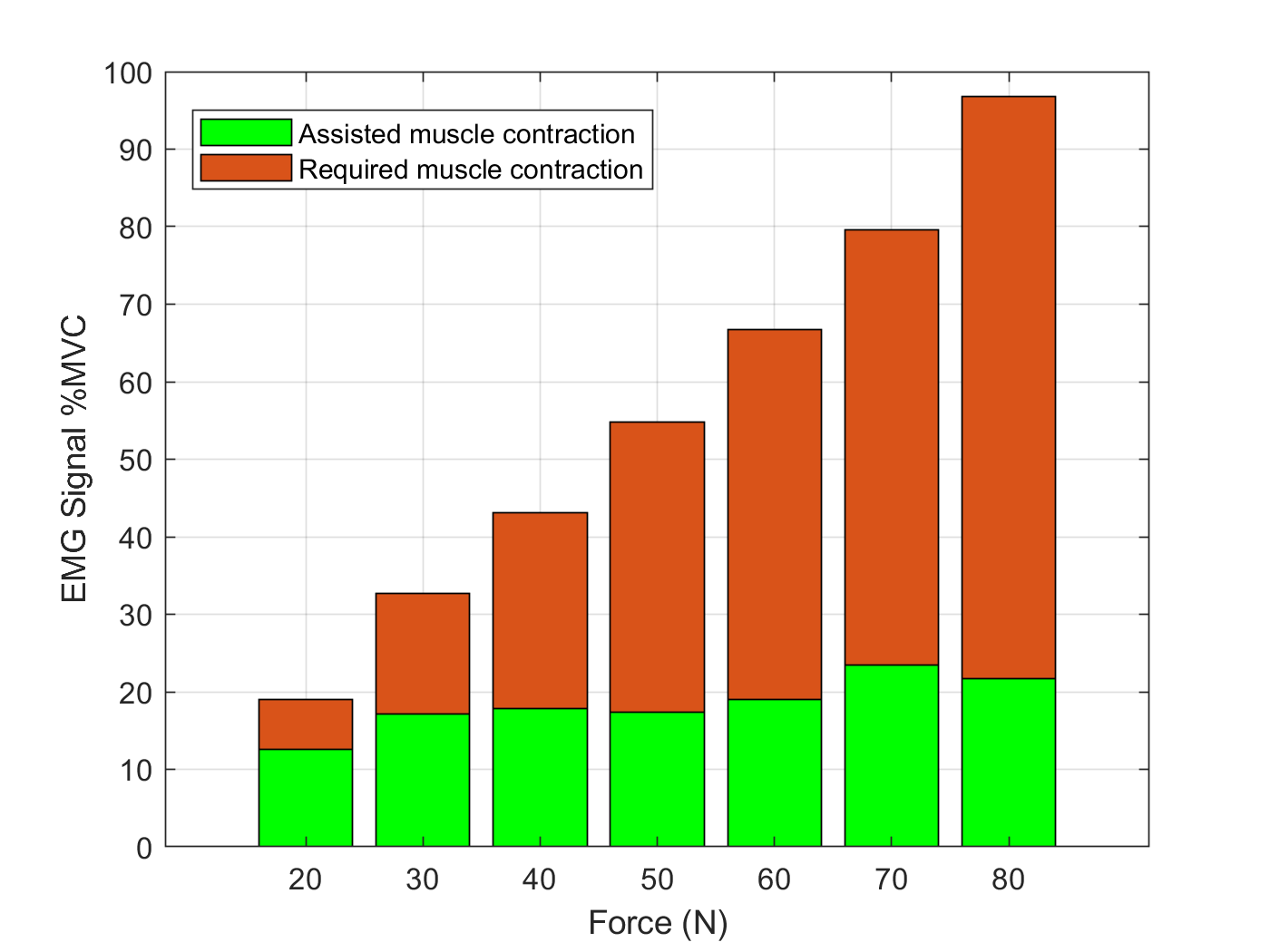}
  \caption{Bar plot that shows muscle contraction assisted by the glove (green), and the muscle contraction needed for the task (red), for different force intensity of tasks.}
  \label{bar}
\end{figure}

In our latest study, we propose a novel method to evaluate wearable devices with healthy participants using EMG sensing. We then apply this method to assess the effectiveness of our soft robotic assistive prototype glove. Our results demonstrate that our exoskeleton significantly improves participants' grasping ability. We record EMG signals as participants perform a task involving varied grasp force applied to an object with and without the assistance of the robotic system. The benefits of the assistive device are quantified using linear mixed-effects models. As our work is currently under review, we provide only a preview of the results in Figure \ref{bar}.

To evaluate the glove's impact, we analyze the average assistance provided to participants, alongside standard deviations, normalized relative to each participant's maximum voluntary contraction (MVC). Assistance is computed by subtracting the RMS EMG signals recorded with the assistive glove from those recorded without it. Figure \ref{bar} illustrates the muscle contraction required for different grasping task intensities (red) and the corresponding reduction in muscle contraction achieved with the assistive device (green). The mean assistance values for all 20 participants' fingers average around 18\% of MVC. Notably, the assistive glove consistently delivers assistance across various task intensities.

These results indicate that the soft robotic assistive glove effectively supplies the majority of the required force for low-force grasping tasks. In a 20N pinch grasp, the glove reduces approximately 70\% of the workload on the fingers, with some instances where the robotic device compensates for the entire workload (100\%) for certain participants. It is observed that when activated, the assistive glove maintains grasp force assistance ranging from 15 to 23N (depending on hand size). 



\section{Conclusions}

 The elastic bands integrated using the ruffles technique proved to be effective in enhancing the performance of the soft robotic structures. In the actuator application, the elastic bands greatly increased the bending capability and force capability of the structure. These findings demonstrate the potential of using elastic bands and textile techniques in soft robotics to create more efficient and adaptable structures.


 Given the assistive glove's capability to apply force exceeding 20N on the fingers, our anticipation for more significant results in the user study was reasonable. However, the outcomes were not as expected, mainly due to the inherent nature of wearable devices, which often elevate EMG readings regardless of actual muscle effort. This phenomenon stems from the "unnatural" haptic sensation induced by the device. Analyzing this behavior is crucial for gaining deeper insights into the functionality of the assistive device.

To delve further into this matter, participants were instructed to fully relax their hand on the test rig while the assistive device was activated. This approach allowed us to isolate the signal increase caused solely by the device activation. Our experiments revealed a mean change of 6.5\% of MVC in EMG signals upon device activation. For some participants, this change was as minimal as 0.0008\% of MVC, suggesting that the EMG alteration cannot be attributed to the static charge of the robotic system. Rather, it reflects the muscle contraction response of the participant to the activation of the assistive glove. Consequently, we can deduce that the assistance provided by the robotic glove exceeds the reported values by approximately 6.5\% of MVC. For an extended version of this two page abstract please visit \cite{10608402}.

\bibliographystyle{IEEEtran}
\bibliography{references.bib}

\end{document}